\documentclass{article}
\usepackage{ijcai23}

\usepackage{amssymb}
\usepackage{times}
\usepackage{soul}
\usepackage{url}
\usepackage{bm}
\usepackage{amsfonts}
\usepackage{adjustbox}
\usepackage{wrapfig}
\usepackage[colorlinks,linkcolor=blue,citecolor={blue!60!black}]{hyperref}
\usepackage[utf8]{inputenc}
\usepackage[small]{caption}
\usepackage{graphicx}
\usepackage{amsmath}
\usepackage{amsthm}
\usepackage{booktabs}
\usepackage{algorithm}
\usepackage{algorithmic}
\usepackage[switch]{lineno}
\usepackage{comment}
\usepackage{multirow}
\usepackage{multicol}
\usepackage{mathtools}
\usepackage{xcolor}
\usepackage{subfigure}
\usepackage{nicefrac}
\usepackage{makecell}

\newcommand{\fig}[1]{Fig.~\ref{#1}}
\newcommand{\eq}[1]{Eq.~(\ref{#1})}

\newcommand{\se}[1]{Section~\ref{#1}}

\newcommand*{\dif}{\mathop{}\!\mathrm{d}}

\newcommand{\citet}[1]{\citeauthor{#1} [\citeyear{#1}]}

\title{Diffusion Models for Reinforcement Learning: A Survey}

\author{
Zhengbang Zhu$^1$
\and
Hanye Zhao$^1$\and
Haoran He$^1$\and
Yichao Zhong$^1$\and\\
Shenyu Zhang$^1$\and
Haoquan Guo$^2$\and
Tingting Chen$^2$\and
Weinan Zhang$^1$\footnote{Corresponding author.}
\affiliations
$^1$Shanghai Jiao Tong University\\
$^2$Shanghai Marine Electronic Equipment Research Institute
\emails
\{zhengbangzhu, wnzhang\}@sjtu.edu.cn
}

\begin{document}

\maketitle

\begin{abstract}

Diffusion models surpass previous generative models in sample quality and training stability. 
Recent works have shown the advantages of diffusion models in improving reinforcement learning (RL) solutions.
This survey aims to provide an overview of this emerging field and hopes to inspire new avenues of research. 
First, we examine several challenges encountered by RL algorithms.
Then, we present a taxonomy of existing methods based on the roles of diffusion models in RL and explore how the preceding challenges are addressed.
We further outline successful applications of diffusion models in various RL-related tasks.
Finally, we conclude the survey and offer insights into future research directions.
We are actively maintaining a GitHub repository for papers and other related resources in utilizing diffusion models in RL
\footnote{\url{https://github.com/apexrl/Diff4RLSurvey}}.

\end{abstract}

\section{Introduction}
Diffusion models have emerged as a powerful class of generative models, garnering significant attention in recent years.
These models employ a denoising framework that can effectively reverse a multistep noising process to generate new data~\cite{song2021scorebased}.
In contrast to earlier generative models such as Variational Autoencoders (VAE)~\cite{kingma2013auto} and Generative Adversarial Networks (GAN)~\cite{goodfellow2014generative}, diffusion models exhibit superior capabilities in generating high-quality samples and demonstrate enhanced training stability.
Consequently, they have made remarkable strides and achieved substantial success in diverse domains including computer vision~\cite{ho2020denoising,lugmayr2022repaint,luo2021diffusion}, natural language processing~\cite{austin2021structured,li2022diffusion}, audio generation~\cite{lee2021nu,kong2020diffwave}, and drug discovery~\cite{xu2022geodiff,schneuing2022structure}, \textit{etc.}

Reinforcement learning (RL)~\cite{sutton2018reinforcement} focuses on training agents to solve sequential decision-making tasks by maximizing cumulative rewards.
While RL has achieved remarkable successes in various domains~\cite{kober2013reinforcement,kiran2021deep}, there are some long-standing challenges.
Specifically, 
despite the considerable attention garnered by offline RL for overcoming low sample efficiency issue in online RL~\cite{kumar2020conservative,fujimoto2021minimalist}, conventional Gaussian policies may fail to fit the datasets with complex distributions for their \textit{restricted expressiveness}.
Meanwhile, although experience replay is used to improve sample efficiency~\cite{mnih2013playing}, there is still \textit{data scarcity} problem in environments with high-dimensional state spaces and complex interaction patterns.
A common usage of learned dynamic models in model-based RL is planning in them~\cite{nagabandi2018neural,schrittwieser2020mastering,zhu2021mapgo}, but the per-step autoregressive planning approaches suffer from the \textit{compounding error} problem~\cite{xiao2019learning}.
An ideal RL algorithm should be able to learn a single policy to perform multiple tasks and generalize to new environments~\cite{vithayathil2020survey,beck2023survey}. However, existing works still struggle in \textit{multitask generalizations}.

Recently, there has been a series of works applying diffusion models in  sequential decision-making tasks, with a particular focus on offline RL.
As a representative work, Diffuser~\cite{Janner2022PlanningWD} fits a diffusion model for trajectory generation on the offline dataset, and plans desired future trajectories by guided sampling. There have been many following works where diffusion models behave as different modules in the RL pipeline, \textit{e.g.}, replacing conventional Gaussian policies~\cite{wang2023diffusion}, augmenting experience dataset~\cite{lu2023synthetic}, extracting latent skills~\cite{venkatraman2023reasoning}, among others.  
We also observe that planning and decision-making algorithms facilitated by diffusion models perform well in broader applications such as multitask RL~\cite{he2023diffusion}, imitation learning~\cite{hegde2023generating}, and trajectory generation~\cite{zhang2022motiondiffuse}.
More importantly, diffusion models have already shed light on resolving those long-standing challenges in RL owing to their powerful and flexible distributional modeling ability.

This survey centers its attention on the utilization of diffusion models in RL, with additional consideration given to methods incorporating diffusion models in the contexts of trajectory generation and imitation learning, primarily due to the evident interrelations between these fields.
\se{sec:challenge} elaborates on the aforementioned RL challenges, and discusses how diffusion models can help solve each challenge. 
\se{sec:diffusion} provides a background on the foundations of diffusion models and also covers two class of methods that are particularly important in RL-related applications: guided sampling and fast sampling.
\se{sec:role} illustrates what roles diffusion models play in RL among existing works.
\se{sec:application} discusses the contribution of diffusion models on different RL-related applications.
\se{sec:summary} summarizes the survey with a discussion on emerging new topics.

\begin{figure*}[ht]
\centering
\includegraphics[width=0.95\linewidth]{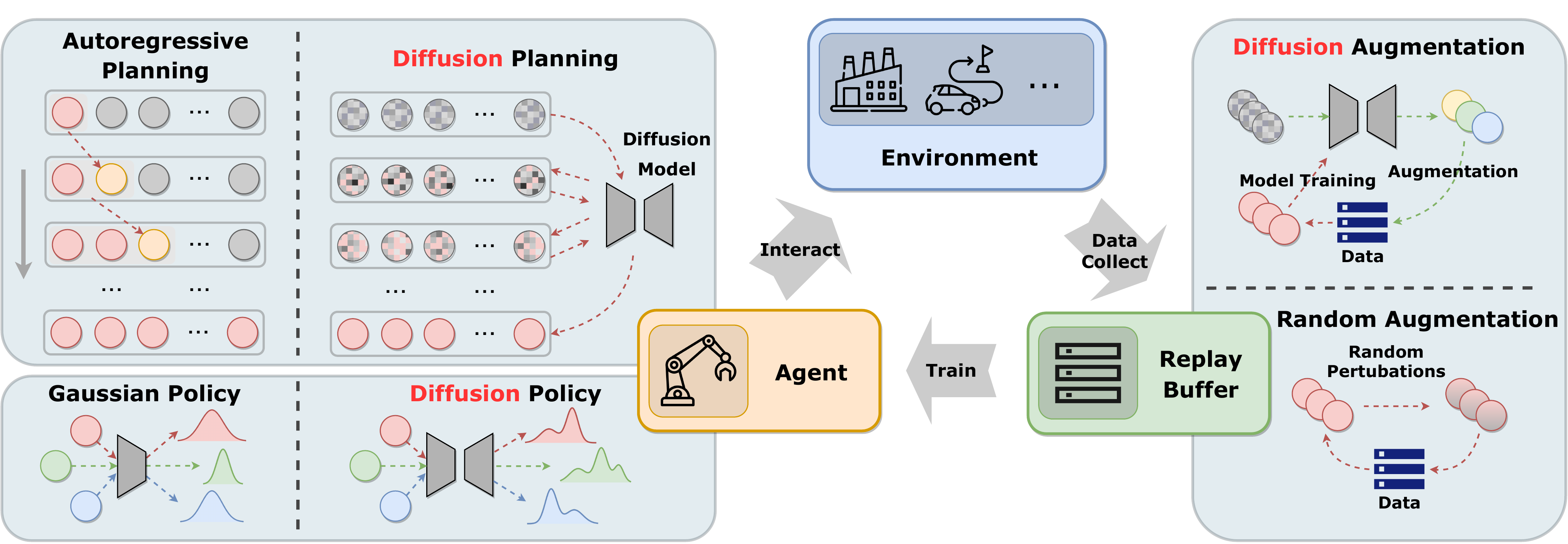}
\caption{
An illustration of how diffusion models play a different role in the classic Agent-Environment-Buffer cycle compared to previous solutions. 
(1) When used as a planner, diffusion models optimize the whole trajectory at each denoising step, whereas the autoregressive models generate the next-step output only based on previously planned partial subsequences.
(2) When used as a policy, diffusion models can model arbitrary action distributions, whereas Gaussian policies can only fit the possibly diversified dataset distribution with unimodal distributions.
(3) When used as a data synthesizer, diffusion models augment the dataset with generated data sampled from the learned dataset distribution, whereas augmentation with random perturbations might generate samples that deviate from data samples.
}
\label{fig:main}
\end{figure*}

\section{Challenges in Reinforcement Learning}
\label{sec:challenge}

In this section, we list four challenges in RL algorithms and briefly discuss why diffusion models can address them.

\subsection{Restricted Expressiveness in Offline Learning}
\label{sec:ch-express}
Due to the low-sample efficiency, online RL~\cite{sutton2018reinforcement} is challenging to be applied in real-world scenarios.
Offline RL~\cite{fujimoto2019off,kumar2020conservative} learns optimal policies from pre-collected datasets without environmental interaction, which can significantly improve sample efficiency.
Directly applying off-policy RL to offline learning causes severe extrapolation errors~\cite{fujimoto2019off}.
Existing works penalize value predictions on out-of-distribution samples~\cite{kumar2020conservative} or limit the learning policy to be close to the data collecting policy~\cite{kostrikov2021offline}.
However, penalties imposed on the value function may result in an overly conservative policy~\cite{lyu2022mildly};
when imposing policy constraints on commonly used unimodal Gaussian parameterization, the restricted expressiveness makes it difficult to fit the possibly diverse dataset.
Reinforcement learning via supervised learning framework (RvS)~\cite{schmidhuber2019reinforcement} is another important paradigm in offline RL, which eliminates Q-learning thus free of extrapolation errors.
RvS learns a policy conditioned on the observed returns via supervised learning and then conditions it on a high return to generate desired behaviors~\cite{chen2021decision}.
Similar to policy constraining, RvS requires fitting the entire dataset.
Therefore, the expressiveness of parameterized policies also matters in RvS.
Diffusion models have the capability to represent any normalizable distribution~\cite{neal2011mcmc}, with the potential to effectively improve the performance of policy constraining and RvS algorithms on complex datasets.

\subsection{Data Scarcity in Experience Replay}
Off-policy and offline RL methods use different levels of experience replay to improve sample efficiency.
Note that experience replay in some literature only refers to data reuse in off-policy RL. Here, the term broadly refers to updating the current model with rollout data from other policies.
Although all previous experiences can be used for policy learning in off-policy RL, the limitation of simulation speed and the potentially huge state and action spaces may still hinder policy optimization.
In offline RL, policy learning is more limited by the quality and coverage of the dataset as no further interactions are allowed.
Inspired by the success of data augmentation in computer vision, some works adopt similar augmentation in RL to reduce data scarcity.
RAD~\cite{laskin2020reinforcement} uses image augmentation such as random cropping or rotation to improve learning efficiency in vision-based RL.
\citet{imre2021investigation} and \citet{cho2022s2p} use generative models, VAE~\cite{kingma2013auto} and GAN~\cite{goodfellow2014generative}, to augment the real dataset with generated synthetic data. 
However, existing works either lack fidelity when using random augmentation or are limited to simple environments due to insufficient modeling ability of particular generative models, making them difficult to be applied to more complex tasks.
Diffusion models have demonstrated notable performances surpassing previous generative models in high-resolution image synthesis~\cite{ho2020denoising}. 
When applied to RL data, diffusion models are better suited for enhancing complex interactions.

\subsection{Compounding Error in Model-based Planning}
MBRL~\cite{luo2022survey} fits a dynamic model from online rollout data or offline datasets to facilitate decision-making.
Common dynamic models mimic single-step state transitions and rewards in the dataset. 
When predicting with a neural dynamic model, there could be single-step errors due to the limited data support and stochastic environment transitions.
Cumulative single-step errors can make planned states deviate from the dataset distribution, which causes the compounding error problem when using the model for multistep planning~\cite{xiao2019learning}.
In contrast, diffusion models with powerful modeling ability of joint distributions can operate on the trajectory level and plan for multiple time steps simultaneously, improving temporal consistency and reducing compounding errors.

\subsection{Generalization in Multitask Learning}
Normal RL algorithms lack generalization abilities at the task level~\cite{beck2023survey}.
Even in the same environment, changing the reward function requires retraining a policy from scratch. 
Existing online multitask RL~\cite{liu2021conflict} works attempt to learn the same policy in different task environments, suffering from conflicting gradients across multiple tasks and low sample efficiency due to pure online learning. 
Recently, it has been a popular research direction to train a high-capacity model on multitask offline datasets and then deploy it on new tasks with or without online fine-tuning~\cite{taiga2022investigating}.
Transformer-based pre-training decision models like Gato~\cite{reed2022generalist} excel at multitask policy learning.
However, they typically require high-quality datasets, large parameter sizes, and high training and inference costs.
In multitask RL, designing an algorithm that can fit mixed-quality multitask datasets and generalize to new tasks emerges as a vital issue.
As a powerful class of generative models, diffusion models can deal with multimodal distributions in multitask datasets, and adapt to new tasks by estimating the task distribution.

\section{Foundations of Diffusion Models}
\label{sec:diffusion}

We introduce the foundations of diffusion models, including two prominent formulations and several sampling techniques that are particularly important in RL tasks.
 
\subsection{Denoising Diffusion Probabilistic Model}
Assuming that the real data $x^0$ are sampled from an underlying distribution $q(x^0)$, DDPM~\cite{ho2020denoising} utilizes a parameterized diffusion process, represented as $p_\theta(x^0)=\int p(x^T)\prod_{t=1}^T p_\theta(x^{t-1}|x^t)\dif x^{1:T}$, to model how the pure noise $x^T=\mathcal{N}(\mathbf{0},\mathbf{I})$ is denoised into real data $x^0$. 
Each step of the diffusion process is represented by $x^t$, with $T$ indicating the total number of steps. Note that both the diffusion process and RL involve time steps; thus, we denote diffusion steps as superscripts and RL time steps as subscripts. The sequence $x^{T:0}$ is defined as a Markov chain with learned Gaussian transitions characterized by $p_\theta(x^{t-1}|x^t)=\mathcal{N}(\mu_\theta(x^t,t),\Sigma(x^t,t))$.
If the process is reversed as $x^{0:T}$, each step is defined by the forward transition $q(x^t|x^{t-1})$, which is formulated as adding Gaussian noise to the data according to a variance schedule $\beta^{1:T}$:
\begin{equation}
    x^t=\sqrt{\alpha^t}x^{t-1}+\sqrt{1-\alpha^t}\epsilon^t ~,\label{eq:xtxt-1}
\end{equation}
where $\alpha^t=1-\beta^t$, $ \epsilon^t\sim\mathcal{N}(\mathbf{0},\mathbf{I})$.
From \eq{eq:xtxt-1}, we can derive a direct mapping from $x^0$ to $x^t$:
\begin{equation*}
    x^t = \sqrt{\bar{\alpha}^t}x^0+\sqrt{1-\bar{\alpha}^t}{\epsilon(x^t,t)} ~,
\end{equation*}
where $\bar{\alpha}^t=\prod_1^t\alpha^i$.
From Bayes theorem and relation between $x^t$ and $x^0$, we have
\begin{equation}
\label{eq:ddpmmu}
q(x^{t-1}|x^t, x^0)
=\mathcal{N}(\frac1{\sqrt{{\alpha}^t}}(x^t-\frac{\beta^t}{\sqrt{1-\bar{\alpha}^t}}\epsilon(x^t,t)), \beta^t\mathbf{I})~.
\end{equation}
\eq{eq:ddpmmu} allows us to sample $x^T$ from Gaussian noise and denoise step by step until we obtain $x^0$.
However, the noise $\epsilon(x^t,t)$ is unknown. To address this, a parameterized network $\epsilon_\theta$ is employed to predict the noise. 
\citet{ho2020denoising} propose the following simplified loss function for learning $\epsilon_\theta$, which is a weighted version of the evidence lower bound (ELBO):
\begin{equation}
    \mathcal{L}(\theta)= \mathbb{E}_{x^0,\epsilon,t}[\|\epsilon-\epsilon_\theta(\sqrt{\bar{\alpha}^t}x^0+\sqrt{1-\bar{\alpha}^t}\epsilon,t)\|^2]~,
    \label{eq:ddpmloss}
\end{equation}
where $\epsilon$ is sampled from $\mathcal{N}(\mathbf{0},\mathbf{I})$.

\subsection{Score-based Generative Models}
\citet{song2021scorebased} extend DDPM to continuous-time diffusion processes, and the sequence $x^0,x^1,\ldots,x^T$ is replaced with a continuous function $x^t, t\in[0,T]$. The forwarding process can be described as a Stochastic Differential Equation (SDE):
\begin{equation*}
    \dif x=f(x,t)\dif t+g(t)\dif w~,
\end{equation*}
where $f(x,t)$ and $g(t)$ are pre-defined functions, and $\dif w$ is the Brownian motion. 
According to Langevin dynamics, the reverse of the forwarding process is described by a reverse-time SDE:
\begin{equation*}
    \dif x=[f(x,t)-g^2(t)\nabla_{x}\log p_t(x)]\dif t+g(t)\dif \bar{w} ~,\label{eq:DenoiseSDE}
\end{equation*}
where $\bar{w}$ is the reverse Brown motion, $p_t(x)$ is the probability density of $x^t$, and $s(x)=\nabla_{x}\log p_t(x)$ is called the score function of $p_t(x)$.
In practice, a parameterized score model $s_\theta$ is adopted to estimate the score function, 
which can be trained by minimizing
\begin{equation*}
    \mathcal{L}(\theta)=\mathbb{E}_{x^0,t,x^t}\left[\|s_\theta(x^t, t)-\nabla_{x^t}\log p(x^t|x^0)\|_2^2\right]~.
\end{equation*}

\subsection{Guided Sampling Methods}
Guided sampling from diffusion models considers sampling from the conditional distribution $p(x|y)$, where $y$ is the desired attribute of generated samples.
Two main categories of guidance are classifier-guidance and classifier-free guidance.

\begin{figure*}[t]
    \centering
    \includegraphics[width=1.0\linewidth]{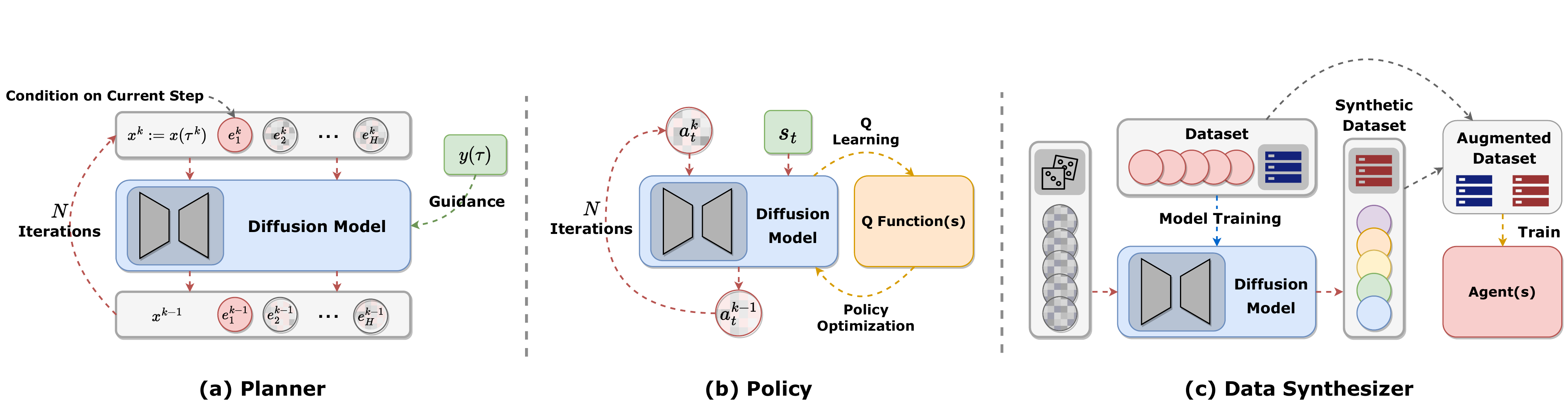}
    \caption{Different roles of diffusion models in RL. (a) Diffusion models as the planner. The sampling target is a part of trajectories whose components may vary from specific tasks. (b) Diffusion models as the policy. The sampling target is the action conditioned on the state, usually guided by the Q-function via policy gradient-style guidance or directly subtracting it from the training objective. (c) Diffusion models as the data synthesizer. The sampling target is also the trajectory, and both real and synthetic data are used for downstream policy improvement.
    For better visualizations, we omit the arrows for $N$ denoising iterations in (c) and only show generated synthetic data from randomly sampled noise.
    Note that there are other roles that are less explored, and we introduce them in \se{sec:other-roles}.}
    \vspace{-12pt}
    \label{fig:roles}
\end{figure*}

\paragraph{Classifier guidance.}
Classifier guided sampling relies on a differentiable classifier model $p_\phi(y|x)$. Specifically, since the guidance needs to be performed on each denoising step, the classifier model $p(y|x^t)$ is trained on noisy samples of $x$ and corresponding attribute $y$. 
The conditional reverse process can be written as
\begin{equation}
    p_{\theta,\phi}(x^{t-1}|x^{t},y)=Z p_\theta(x^{t-1}|x^{t})p_\phi(y|x^{t-1})~,\label{eq:classifierguidance}
\end{equation}
where $Z$ is the normalization factor. \citet{dhariwal2021diffusion} approximate \eq{eq:classifierguidance} by another Gaussian distribution:
\begin{equation}\label{eq:classifier-gaussian}
    p_{\theta,\phi}(x^{t-1}|x^{t},y)=\mathcal{N}(\mu^t+w\Sigma^t g,\Sigma^t)~,
\end{equation}
where $g=\nabla_{x^t}\log p_\phi(y|x^t)|_{x^t=\mu^t}$ and $w$ is the guidance scale to control the strength of conditions. $\mu^t$ and $\Sigma^t$ are the mean and the covariance matrix in \eq{eq:ddpmmu}, respectively.

\paragraph{Classifier-free guidance.}
Classifier-free sampling relies on an extra conditional noise model $\epsilon_\theta(x^t,y,t)$.
In practice, the conditional and unconditional models share the same set of parameters, and the unconditional model is represented by setting $y$ as a dummy value $\varnothing$.
\citet{ho2022classifierfree} state that the noise learning target in \eq{eq:ddpmloss} is a scaled score function of $p(x^t)$, \textit{i.e.}, $\epsilon(x^t,t)=-{\sigma^t}\nabla_{x^t}\log p(x^t)$ and $\sigma^t=\sqrt{\beta_t}$.
By using Bayes theorem, we have
\begin{equation*}
    \nabla_{x^t}\log p(y|x^t) = -1/\sigma^t(\epsilon(x^t,y,t)-\epsilon(x^t,t))~.
\end{equation*}
According to \eq{eq:classifier-gaussian}, we can derive the guided noise predictor as $\bar{\epsilon}_\theta(x^t,y,t)=\epsilon_\theta(x^t,t)- w\sigma^t \nabla_{x^t}\log p(y|x^t)$.
Replacing the score function with the noise model predictions, the noise used in classifier-guided sampling can be written as
\begin{equation*}
    \hat{\epsilon}_w(x^t,y,t)=w\epsilon_\theta(x^t,y,t)+(1-w)\epsilon_\theta(x^t,t)~.
\end{equation*}

\subsection{Fast Sampling Methods}
Various fast sampling methods are proposed to overcome the prolonged iterative sampling time of diffusion models.
We summarize these methods into two categories: those that do not involve learning and those that require extra learning, and describe representative works in each category.

\paragraph{Learning-free methods.}
DDIM~\cite{song2022denoising}
is one of the seminal works on sampling acceleration. It extends DDPM to a non-Markovian formulation by learning another Markov chain $q_\theta(x^{t-1}|x^t,x^0)$. 
Some high-order solvers are proposed for diffusion sampling, such as DPM-solver~\cite{lu2022dpmsolver}, which provides an excellent trade-off between sample quality and sampling speed. With DDIM as its first-order version, DPM-solver boosts the efficiency of solving PF-ODE, outperforming common numerical ODE solvers. 

\paragraph{Learning-based methods.}
Learning-based sampling methods require extra training to obtain a higher sampling efficiency at a slight expense of sampling quality. A recent work, Truncate Diffusion Probabilistic Model (TDPM)~\cite{zheng2023truncated}, demonstrates that both the noising and denoising process can be early terminated to reduce the iterative steps. 
Moreover, \citet{watson2021learning} propose a strategy to select the best $K$ time steps to maximize the training objective for the DDPMs, which also decreases the denoising steps.

\section{The Roles of Diffusion Models in RL}
\label{sec:role}

\fig{fig:main} illustrates how diffusion models play a different role in RL compared to previous solutions.
Current works applying diffusion models on RL mainly fall into three categories: as planners, as policies, and as data synthesizers.
It is essential to note that we include methods that generate action-only sequences as planners, even though some of the representative works have ``policy" in their names, e.g., Diffusion Policy~\cite{chi2023diffusion}.
Generating multi-step action sequences can be viewed as planning in action space, and the use of diffusion models to ensure temporal consistency is similar to other planning-based diffusion methods.
The following subsections will illustrate overall frameworks and representative papers for each category.

\subsection{Planner}
Planning in RL refers to using a dynamic model to make decisions imaginarily and selecting the appropriate action to maximize cumulative rewards. 
This process usually explores various sequences of actions and states, 
thus improving decisions over a longer horizon. Planning is commonly used in the MBRL framework with a learned dynamic model. However, the planning sequences are usually simulated autoregressively, which may lead to severe compounding errors, especially in the offline setting due to limited data support. 
Diffusion models offer a promising alternative as they can generate multi-step planning sequences simultaneously. 

A general framework of diffusion planners is shown in \fig{fig:roles}(a). Diffusion models are designed to generate clips of the trajectory $\tau=(s_1,a_1,r_1,\ldots,s_H,a_H,r_H)$, denoted as $x(\tau)=(e_1,e_2,\ldots,e_H)$. $H$ is the planning horizon. Here $e_t$ represents the selected elements from $(s_t,a_t,r_t)$, where various choices can be made as $e_t=(s_t,a_t)$~\cite{Janner2022PlanningWD,adaptdiffuser,he2023diffusion}, $e_t=(s_t,a_t,r_t)$~\cite{he2023diffusion,hu2023instructed}, $e_t=s_t$~\cite{decision_diffuser,zhu2023madiff}, or $e_t=a_t$~\cite{chi2023diffusion,li2023crossway}.
RL datasets often contain trajectory data of varying quality. In order to make the diffusion planner generate high rewarded trajectories during evaluations, guided sampling techniques are widely adopted.
The guidance can be either injected in the sampling stage following the classifier guided sampling or in both the training and sampling stages following the classifier-free sampling.

When deploying the trained diffusion planner for online evaluations, fast sampling methods are usually adopted to reduce the inference time.
Besides, to ensure the planned trajectory is congruous with the agent's current state, before each denoising step, the first $h \,(h\geq 1)$ steps of the noisy trajectory are substituted with the $h$ steps of historical states observed by the agent. Here, $h$ is a hyperparameter where a larger $h$ can better handle partially observable and non-Markovian settings but increases the modeling complexity.

\subsection{Policy}
Based on whether they rely on a dynamic model for making decisions, RL algorithms can be categorized into MBRL and model-free RL. Under such classification criteria, using diffusion models as planners is more akin to MBRL, as the generated trajectories encapsulate dynamics information.
Another perspective is that diffusion planner can be seen as a combination of policy and dynamic model~\cite{Janner2022PlanningWD}.
In contrast, using diffusion models as policies focuses on improving existing model-free RL solutions. \se{sec:ch-express} states the main drawbacks of current offline policy learning methods: over-conservatism and lack of expressiveness. Many works use diffusion models as the policy class in model-free RL to tackle these problems.

Diffusion-QL~\cite{wang2023diffusion} first combines the diffusion policy with the Q learning framework and finds that it can perfectly fit on datasets collected by strong multi-modal behavior policies, where previous distance-based policy regularization approaches fail.
Compared with using diffusion models as planners, the diffusion target of the diffusion policy is simply the action given the current state, as shown in \fig{fig:roles}(b). Suppose the noise predictor is $\epsilon_\theta(a^k, k,s)$ parameterized by $\theta$, and the derived action mean is $\mu_\theta(a|s)$. To guide the model sampling actions that can lead to high returns, it is necessary to take $Q(s, a)$ into consideration.
Diffusion-QL includes a weighted Q maximization term into the diffusion training loss as
\begin{equation*}\label{eq:subt_q}
\begin{aligned}
    \mathcal{L}(\theta)&=\mathbb{E}_{k,\epsilon,(s,a)\sim\mathcal{D}}[\|\epsilon - \epsilon_\theta(a^k,s,k)\|_2^2] \\
    &\quad- \frac{\eta}{\mathbb{E}_{(s,a)\sim\mathcal{D}}[Q(s,a)]}\cdot \mathbb{E}_{s\sim\mathcal{D}, a^0\sim\pi_\theta(\cdot|s)}[Q(s,a^0)]~,
\end{aligned}
\end{equation*}
where $\eta$ is a hyperparameter, and $\mathcal{D}$ is the offline dataset.
Some works~\cite{chen2023offline,lu2023contrastive,hansenestruch2023idql,kang2023efficient} construct the policy by (advantage) weighted regression as
\begin{equation*}\label{eq:weighted_reg}
    \pi_\theta(a|s) \propto \mu_\theta(a|s)\exp(\alpha Q(s,a))~,
\end{equation*}
where $\alpha$ is the temperature hyperparameter. Following this, \citet{chen2023offline} decouple the policy learning into behavior learning and action evaluation, which allows more freedom in the choice of guidance. They also propose in-sample planning for Q-learning, avoiding extrapolation errors in previous offline RL methods. CEP~\cite{lu2023contrastive} further extends this framework to sample from the more general energy-guided distribution $p(x)\propto q(x)\exp(-\beta \mathcal{E}(x))$. Here, $\mathcal{E}(x)$ is an energy function, and in the RL setting, it is trained via contrastive learning to match the in-sample softmax of Q functions.
Since there are off-the-shelf Q functions or energy functions after training, some methods use those functions to further improve the sampling quality during evaluations. They first sample multiple candidate actions for a given state and use Q or energy values to perform weighted sampling or just select the best candidate.

Fast reaction is crucial when deploying policies in online environments. Therefore, almost all diffusion policies use smaller diffusion steps during sampling, usually about 15 steps. ODE solvers such as the DPM-solver~\cite{lu2022dpmsolver} are also used to accelerate sampling~\cite{chen2023offline,lu2023contrastive,kang2023efficient,li2023conservatism}.
\citet{kang2023efficient} introduce action approximation, which allows one-step action sampling in the training stage.

\subsection{Data Synthesizer}
In addition to fitting multi-modal distributions, a simple and common use of diffusion models is to create synthetic data, which has been widely applied in computer vision. Therefore, the diffusion model is a natural data synthesizer for RL datasets because data scarcity is a practical concern. 
To ensure consistency of synthetic data to the environment dynamics, previous data augmentation approaches in RL usually add minor perturbations to states and actions~\cite{sinha2021s4rl}. 
In contrast, \fig{fig:roles}(c) illustrates that diffusion models generate diverse and consistent data by learning the data distribution from the entire dataset $\mathcal{D}_{\mathrm{real}}$. 
The diffusion model first learns the parameterized data distribution $\rho_\theta(\tau)$ from the real data $\mathcal{D}_\mathrm{real}$, and generates desired synthetic data by
\begin{equation*}
\mathcal{D}_{\mathrm{syn}} = \{\tau \sim \rho_\theta(\tau)\}~.
\end{equation*}
Then, real and synthetic data are combined together as
\begin{equation*}
    \mathcal{D} = \mathcal{D}_\mathrm{real} \cup \mathcal{D}_\mathrm{syn}~,
\end{equation*}
and $\mathcal{D}$ is used for policy learning.
In online settings, the policy interacts with the environment, collects more real data into $\mathcal{D}_\mathrm{real}$, and updates the diffusion model. As a result, the diffusion model and the policy are updated alternately.

\subsection{Others}
\label{sec:other-roles}
Besides the primary directions discussed above, other ways of improving RL with diffusion models are less explored.
\citet{mazoure2023value} estimate value functions with diffusion models by learning the discounted state occupancy, combined with a learned reward estimator. Then, the value function can be directly computed by definition, where future states are sampled from the diffusion model. \citet{venkatraman2023reasoning} first encode the high-level trajectories into semantically rich latent representations, then apply diffusion models to learn the latent distribution. Conditioning on latent representations improves the capability of Q-functions and policies without significant extrapolation errors.
\citet{rigter2023world} use a diffusion dynamic model and allow an online RL policy to collect synthetic trajectories on it. The interplay of the diffusion dynamic model and RL policy is done by alternating between state denoising and Langevin dynamics of policy actions.

\begin{table*}[hp]
    \centering
    \caption{Summary of papers on diffusion models for RL.}
    \label{tab:main}
    \resizebox{\textwidth}{!}{
    \begin{tabular}{c|c|c|c}
    \toprule
      Model \& Paper & Role of Diffusion Models & Keyword(s) & Guidance \\
    \midrule
      Diffuser~\cite{Janner2022PlanningWD} & \multirow{25}*{Planner} & Offline & Classifier \\
      AdaptDiffuser~\cite{adaptdiffuser} &  & Offline & Classifier \\
      EDGI~\cite{brehmer2023edgi} &  & Offline & Classifier \\
      TCD~\cite{hu2023instructed} &  & Offline & Classifier-free \\
      LatentDiffuser~\cite{li2023efficient} & & Offline & Energy Function \\
      HDMI~\cite{Hier_diffusion} &  & Offline; Hierarchical & Classifier-free \\
      SafeDiffuser~\cite{xiao2023safediffuser} &  & Offline; Safe & None \\
      MADiff~\cite{zhu2023madiff} &  & Offline; Multi-agent & Classifier-free \\
      MTDiff-p~\cite{he2023diffusion} &  & Offline; Multitask & Classifier-free \\
      MetaDiffuser~\cite{ni2023metadiffuser} &  & Offline; Multitask & Classifier-free \\
      Diffusion Policy~\cite{chi2023diffusion} &  & Imitation; Robotics & None \\
      Crossway Diffusion~\cite{li2023crossway} &  & Imitation; Robotics & None \\
      AVDC~\cite{ko2023learning} &  & Imitation; Robotics & None \\
      SkillDiffuser~\cite{liang2023skilldiffuser} & & Imitation; Multitask; Hierarchical & Classifier-free \\
      MLD~\cite{Chen2022ExecutingYC} &  & Trajectory Generation & Classifier-free \\
      MDM~\cite{tevet2022humandiffuse} &  & Trajectory Generation & Classifier-free \\
      UniSim~\cite{yang2023learning} &  & Trajectory Generation & Classifier-free \\
      ReMoDiffuse~\cite{zhang2023remodiffuse} &  & Trajectory Generation & Classifier-free \\
      SinMDM~\cite{raab2023single} &  & Trajectory Generation & None \\
      EquiDiff~\cite{chen2023equidiff} &  & Trajectory Generation & None \\
      MoFusion~\cite{Dabral2022MoFusionAF} &  & Trajectory Generation & None \\
      MotionDiffuse~\cite{zhang2022motiondiffuse} &  & Trajectory Generation & None \\
      MPD~\cite{carvalho2023motion} &  & Trajectory Generation; Robotics & Classifier \\
      MotionDiffuser~\cite{jiang2023motiondiffuser} &  & Trajectory Generation; Multi-agent & Classifier \\
      AlignDiff~\cite{dong2023aligndiff} &  & RLHF & Classifier-free \\
      \midrule
      Diffusion-QL~\cite{wang2023diffusion} & \multirow{16}*{Policy} & Offline & Q-loss \\
      SRDP~\cite{ada2023diffusion} &  & Offline & Q-loss \\
      EDP~\cite{kang2023efficient} &  & Offline & Q-loss \\
      SfBC~\cite{chen2023offline} &  & Offline & Sample \& Reweight \\
      IDQL~\cite{hansenestruch2023idql} &  & Offline & Sample \& Reweight \\
      DiffCPS~\cite{he2023diffcps} &  & Offline & Convex Optimization \\
      CPQL~\cite{chen2023boosting} &  & Offline; Online & Q-loss \\
      CEP~\cite{lu2023contrastive} &  & Offline; Image Synthesis & Energy Function \\
      DOM2~\cite{li2023conservatism} &  & Offline; Multi-agent & Q-loss \\
      NoMaD~\cite{sridhar2023nomad} &  & Imitation; Robotics & None \\
      BESO~\cite{reuss2023goalconditioned} &  & Imitation; Goal-conditioned & Classifier-free \\
     ~\citet{pearce2023imitating} &  & Imitation & Classifier-free \\
     ~\citet{yoneda2023noise} &  & Imitation; Robotics & None \\
      PlayFusion~\cite{chen2023playfusion} &  & Imitation; Robotics & None \\
      XSkill~\cite{xu2023xskill} &  & Imitation; Robotics & None \\
      CoDP~\cite{ng2023diffusion} &  & Human-in-the-loop & None \\
      \midrule
      GenAug~\cite{chen2023genaug} & \multirow{4}*{Data Synthesizer}  & Robotics & None \\
      ROSIE~\cite{yu2023scaling} &  & Robotics & None \\
      SynthER~\cite{lu2023synthetic} &  & Offline; Online & None \\
      MTDiff-s~\cite{he2023diffusion} &  & Offline; Multitask & Classifier-free \\
      \midrule
      LDCQ~\cite{venkatraman2023reasoning} & Latent Representation & Offline & Classifier-free \\
      \midrule
      DVF~\cite{mazoure2023value} & Value Function & Offline & None \\
      \midrule
      PolyGRAD~\cite{rigter2023world} & Dynamic Model & Online & None \\
    \bottomrule
    \end{tabular}}
\end{table*}

\section{Applications of Diffusion Models}
\label{sec:application}
In this section, we conduct a complete review of existing works. We divide them into five groups based on the tasks they are applied: offline RL, online RL, imitation learning, trajectory generation, and data augmentation. For each group, we provide a detailed explanation of how each method uses diffusion models to handle the task.

\subsection{Offline RL}
Offline RL
aims to learn a policy from previously collected datasets without online interaction. Assuming there is a static dataset $\mathcal{D}$ collected by some (unknown) behavior policy $\pi_\beta$, offline RL requires the learning algorithm to derive a policy $\pi(a|s)$ that attains the most cumulative reward:
\begin{equation*}
\label{eq:objective}
\pi^*=\arg\max_{\pi}\mathbb{E}_{\tau\sim p_\pi(\tau)}\Big[\sum_{t=0}^{H}\gamma^t r(s_t,a_t)\Big]~.
\end{equation*}
The fundamental challenge in offline RL is the distributional shift.
This refers to the discrepancy between the dataset distribution used to train the function approximators (\textit{e.g.}, policies and value functions) and the distribution on which the policy is evaluated. 
This mismatch often results in subpar online performance.
High-dimensional and expressive function approximation generally exacerbates this issue.

Several methods use diffusion models to help tackle or avoid the above challenges. \citet{Janner2022PlanningWD} first propose to generate optimal trajectories through iterative denoising with classifier-guided sampling. Subsequent works~\cite{wang2023diffusion,chen2023offline,he2023diffcps,ada2023diffusion,brehmer2023edgi,hansenestruch2023idql} represent the policy as a diffusion model to capture multi-modal distributions and enhance the expressiveness of the policy class, which is beneficial to relieve the approximation error between the cloned behavior policy and true behavior policy. 
\citet{decision_diffuser} alleviate the distribution shift problem by generating state sequences with conditional diffusion models followed by inverse dynamic functions to derive executable actions, which propose a novel approach to use classifier-free guidance with low-temperature sampling to denoise out return-maximizing trajectories.
LatentDiffuser~\cite{li2023efficient} performs diffusion planning over a learned latent space with separate decoders to recover raw trajectories. Benefiting from a more compact planning space, it achieves superior performances on long-horizon and high-dimensional tasks.
In order to improve the generation ability of diffusion models for RL, \citet{lu2023contrastive} propose a new guidance method named contrastive energy prediction and \citet{hu2023instructed} capture more temporal conditions.
By incorporating control-theoretic invariance into the diffusion dynamics, SafeDiffuser \cite{xiao2023safediffuser} guarantees the safe generation of planning trajectories. 
HDMI~\cite{Hier_diffusion} adopts a hierarchical structure to tackle long-horizon decision-making problems, which uses a reward-conditional model to discover sub-goals and a goal-conditional model to generate actions.
\citet{dong2023aligndiff} condition the diffusion planner on diverse behavior attributes and learn from human preferences to generate trajectories that can match user-customized behaviors.
CPQL~\cite{chen2023boosting} leverages consistency models as the policy class for fast training and sampling, while EDP~\cite{kang2023efficient} achieves speed-up during training by using single-step model predictions as action approximations.  Diffusion models are also used as the value function~\cite{mazoure2023value} and representation model~\cite{venkatraman2023reasoning} to facilitate training of normal RL policies.
Recent research has made progress in using diffusion models to improve the performance of policies in multitask and multi-agent offline RL. 

\paragraph{Multitask offline RL.}
Diffusion models are verified to have the potential to address the challenge of multi-task generalization in offline RL. \citet{he2023diffusion} first extend the conditional diffusion model to be capable of solving multitask decision-making problems and synthesizing useful data for downstream tasks. 
LCD~\cite{language-control-diffusion} leverages a hierarchical structure to achieve long-horizon multi-task control. 
MetaDiffuser~\cite{ni2023metadiffuser} demonstrates that incorporating the conditional diffusion model into the context of task inference outperforms previous meta-RL methods. AdaptDiffuser~\cite{adaptdiffuser} combines bootstrapping and diffusion-based generative modeling together to enable the model to adapt to unseen tasks. 

\paragraph{Multi-agent offline RL.}
Using diffusion models in multi-agent RL helps model discrepant behaviors among agents and reduces approximation error. 
MADiff~\cite{zhu2023madiff} uses an attention-based diffusion model to model the complex coordination among behaviors of multiple agents, which is well-suited to learning complex multi-agent interactions. DOM2~\cite{li2023conservatism} incorporates the diffusion model into the policy classes to enhance learning and makes it possible to generalize to shifted environments well.

\subsection{Online RL}
Recently, there have been some works showing that diffusion models can also boost online RL training.
Value estimations in online RL are noisier and change with the current policy, which poses additional challenges on training a multistep diffusion model.
DIPO~\cite{yang2023policy} proposes an action relabeling strategy to perform policy improvement at the data level, bypassing the potentially unstable value-guided training. Actions in the online rollout dataset are updated with gradient ascent, and the diffusion training objective is just supervised learning on the relabeled dataset.
\citet{chen2023boosting} conduct experiments to verify that consistency models with one-step sampling can naturally serve as online RL policies and achieve a balance between exploration and exploitation. 
Instead of using diffusion models as policies, \citet{rigter2023world} build a diffusion dynamic model to generate synthetic trajectories that are consistent with online RL policies.
Applications of diffusion models in online RL are less explored compared to offline settings and merit further investigation.

\subsection{Imitation Learning}
The goal of imitation learning (IL) is to reproduce behavior similar to experts in the environment by extracting knowledge from expert demonstrations. Recently, many works~\cite{hegde2023generating,ng2023diffusion,chen2023playfusion,Kapelyukh2022DALLEBotIW} have demonstrated the efficacy of representing policies as diffusion models to capture multi-modal behavior. \citet{pearce2023imitating} apply diffusion models to imitate human behavior in sequential environments, in which diffusion models are compared with other generative models and viable approaches are developed to improve the quality of behavior sampled from diffusion models. \citet{chi2023diffusion,xian2023chaineddiffuser} generate the robot's behavior via a conditional denoising diffusion process on robot action space. Experiment results show that Diffusion models are good at predicting closed-loop action sequences while guaranteeing temporal consistency~\cite{chi2023diffusion}.
\citet{li2023crossway} improve the models in \citet{chi2023diffusion} by incorporating an auxiliary reconstruction loss on intermediate representations of the reverse diffusion process.
Beneficial from its powerful generation ability, leveraging diffusion models to acquire diverse skills to handle multiple manipulation tasks is promising \cite{chen2023playfusion,mishra2023generative,xu2023xskill,ha2023scaling}.
Diffusion models are already applied to goal-conditioned IL: \citet{reuss2023goalconditioned} use a decoupled score-based diffusion model to learn an expressive goal-conditional policy. In contrast, \citet{sridhar2023nomad} build a unified diffusion policy to solve both goal-directed navigation and goal-agnostic exploration problems.
\citet{liang2023skilldiffuser} adopt a hierarchical structure where the high-level skills are determined by the current visual observation and language instructions. 
Therefore the low-level skill-conditioned planner can satisfy the user-specified multitask instructions.

\subsection{Trajectory Generation}
Trajectory generation aims to produce a dynamically feasible path that satisfies a set of constraints. We focus on using diffusion models to generate human pose and robot interaction sequences, which are more related to the decision-making scenario. 
Many works~\cite{zhang2022motiondiffuse,jiang2023motiondiffuser,tevet2022humandiffuse,zhang2023remodiffuse,Chen2022ExecutingYC,Dabral2022MoFusionAF} have remarked that the conditional diffusion models perform better than traditional methods which use GAN or Transformer. Employing a denoising-diffusion-based framework, they achieve diverse and fine-grained motion generation with various conditioning contexts \cite{chen2023equidiff,carvalho2023motion}.
Recent works~\cite{du2023video,ko2023learning,du2023learning} harness diffusion models to synthesize a set of future frames depicting its planned actions in the future, after which control actions are extracted from the generated video. 
This approach makes it possible to train policies solely on RGB videos and deploy learned policies to various robotic tasks with varying dynamics~\cite{black2023zeroshot,gao2023pretrained}. 
UniSim~\cite{yang2023learning} uses diffusion models to build a universal simulator of real-world interaction by learning through combined diverse datasets. It can be used to train both high-level vision-language planners and low-level RL policies, demonstrating powerful emulation ability.

\subsection{Data Augmentation}
Since diffusion models perform well in learning over multimodal or even noisy distributions, they can model original data distribution precisely. What is more, they are capable of generating diverse data points to expand original distribution while maintaining dynamic accuracy. 
Recent works~\cite{yu2023scaling,chen2023genaug} consider augmenting the observations of robotic control using a text-guided diffusion model while maintaining the same action. 
The recently proposed SynthER~\cite{lu2023synthetic} and MTDiff-s~\cite{he2023diffusion} generate complete transitions of trained tasks via a diffusion model. 
\citet{lu2023synthetic} directly train the diffusion model from the offline dataset or the online replay buffer and then generate samples for policy improvement. Analysis shows that both diversity and accuracy of data generated by diffusion models are higher than those generated by prior data augmentation methods.
\citet{he2023diffusion} deploy diffusion synthesizer on multi-task offline datasets and achieve better performance than that on single-task datasets. They claim that fitting on multiple tasks may enable implicit knowledge sharing across tasks, which also benefits from the multi-modal property of diffusion models. 
These works demonstrate that diffusion augmentation can bring significant improvement in policy learning for IL, online RL, and offline RL.

\section{Summary and Future Prospects}
\label{sec:summary}

This survey offers a comprehensive overview of contemporary research endeavors concerning the application of diffusion models in the realm of RL.
According to the roles played by diffusion models, we categorize existing methods into using diffusion models as planners, policies, data synthesizers, and less popular roles such as value functions, representation models, \textit{etc.} 
By comparing each class of methods to traditional solutions, we can see how the diffusion model addresses some of the longstanding challenges in RL, \textit{i.e.}, restricted expressiveness, data scarcity, compounding error, and multitask generalization. 
It is worth emphasizing that the incorporation of diffusion models into RL remains an emerging field, and there are many research topics worth exploring.
Here, we outline four prospective research directions, namely, generative simulation, integrating safety constraints, retrieval-augmented generation, and composing different skills.

\paragraph{Generative simulation.}
As shown in \fig{fig:main}, existing works use diffusion models to overcome certain limitations of previous solutions in both the agent and buffer parts.
However, there has been a scarcity of research focused on using diffusion modeling to improve the environment. Gen2Sim~\cite{katara2023gen2sim} uses text-to-image diffusion models to generate diverse objects in simulation environments, where RL policies are trained to learn robot manipulation skills.
Besides generating objects in the scene, diffusion models have the potential for broader applications in generative simulation, such as the generation of various possible dynamics functions, reward functions, or opponents in multi-agent learning.

\paragraph{Integrating safety constraints.}
Making decisions in real tasks often necessitates compliance with various safety constraints.
Several safe RL methods transform a constrained RL problem to its unconstrained equivalent~\cite{achiam2017constrained}, which is then solved by generic RL algorithms. 
Policies acquired through these methods remain tailored to the specific constraint threshold specified during training.
A recent research~\cite{liu2023constrained} has extended the applicability of decision transformers to the context of safety-constrained settings, thereby enabling a single model to adapt to diverse thresholds by adjusting the input cost-to-go.
Similarly, diffusion models have the potential to be deployed in safe RL by viewing safety constraints as sampling conditions. \citet{decision_diffuser} demonstrate that a diffusion-based planner can combine different movement skills to produce new behaviors.
Also, classifier-guided sampling can include new conditions by learning additional classifiers, while the parameters of the diffusion model remain unchanged~\cite{dhariwal2021diffusion}.
This makes the diffusion model promising for scenarios with new safety requirements after model training.

\paragraph{Retrieval-enhanced generation.}
Retrieval techniques are employed in various domains such as recommender systems~\cite{qin2020user} and large language models~\cite{kandpal2023large} to enhance the model capacity and handle long-tail distributed datasets.
Some works utilize retrieved data to boost text-to-image and text-to-motion diffusion generation~\cite{sheynin2022knn,zhang2023remodiffuse}, promoting better coverage of uncommon condition signals. 
During online interactions, RL agents may also encounter states that are rare in the training dataset.
By retrieving relevant states as model inputs, the performance of diffusion-based decision models can be improved in these states.
Also, if the retrieval dataset is constantly updated, diffusion models have the potential to generate new behaviors without retraining.

\paragraph{Composing different skills.}
From the perspective of skill-based RL, it is promising to break down complex tasks into smaller, more manageable sub-skills. Diffusion models excel in modeling multi-modal distributions, and since multiple sub-skills can be viewed as distinct modes within the distribution of possible behaviors, they offer a natural fit for this task. Combining with classifier guidance or classifier-free guidance, diffusion models are possible to generate proper skills to complete the facing task. Experiments in offline RL also suggest that diffusion models can share knowledge across skills and combine them up~\cite{decision_diffuser}, thus having the potential for zero-shot adaptation or continuous RL by composing different skills.

\appendix



\section*{Acknowledgments}
The work is partially supported by National Key R\&D Program of China (2022ZD0114804) and National Natural Science Foundation of China (62076161). We thank Minghuan Liu, Xihuai Wang, Jingxiao Chen and Mingcheng Chen for valuable suggestions and discussions.


\bibliographystyle{named}
\bibliography{ijcai23}

\begin{thebibliography}{}

\bibitem[\protect\citeauthoryear{Achiam \bgroup \em et al.\egroup }{2017}]{achiam2017constrained}
Joshua Achiam, David Held, Aviv Tamar, and Pieter Abbeel.
\newblock Constrained policy optimization.
\newblock In {\em International conference on machine learning}, pages 22--31. PMLR, 2017.

\bibitem[\protect\citeauthoryear{Ada \bgroup \em et al.\egroup }{2023}]{ada2023diffusion}
Suzan~Ece Ada, Erhan Oztop, and Emre Ugur.
\newblock Diffusion policies for out-of-distribution generalization in offline reinforcement learning, 2023.

\bibitem[\protect\citeauthoryear{Ajay \bgroup \em et al.\egroup }{2023}]{decision_diffuser}
Anurag Ajay, Yilun Du, Abhi Gupta, Joshua~B. Tenenbaum, T.~Jaakkola, and Pulkit Agrawal.
\newblock Is conditional generative modeling all you need for decision-making?
\newblock 2023.

\bibitem[\protect\citeauthoryear{Austin \bgroup \em et al.\egroup }{2021}]{austin2021structured}
Jacob Austin, Daniel~D Johnson, Jonathan Ho, Daniel Tarlow, and Rianne Van Den~Berg.
\newblock Structured denoising diffusion models in discrete state-spaces.
\newblock {\em Advances in Neural Information Processing Systems}, 34:17981--17993, 2021.

\bibitem[\protect\citeauthoryear{Beck \bgroup \em et al.\egroup }{2023}]{beck2023survey}
Jacob Beck, Risto Vuorio, Evan~Zheran Liu, Zheng Xiong, Luisa Zintgraf, Chelsea Finn, and Shimon Whiteson.
\newblock A survey of meta-reinforcement learning.
\newblock {\em arXiv preprint arXiv:2301.08028}, 2023.

\bibitem[\protect\citeauthoryear{Black \bgroup \em et al.\egroup }{2023}]{black2023zeroshot}
Kevin Black, Mitsuhiko Nakamoto, Pranav Atreya, Homer Walke, Chelsea Finn, Aviral Kumar, and Sergey Levine.
\newblock Zero-shot robotic manipulation with pretrained image-editing diffusion models, 2023.

\bibitem[\protect\citeauthoryear{Brehmer \bgroup \em et al.\egroup }{2023}]{brehmer2023edgi}
Johann Brehmer, Joey Bose, Pim de~Haan, and Taco Cohen.
\newblock Edgi: Equivariant diffusion for planning with embodied agents, 2023.

\bibitem[\protect\citeauthoryear{Carvalho \bgroup \em et al.\egroup }{2023}]{carvalho2023motion}
Joao Carvalho, An~T. Le, Mark Baierl, Dorothea Koert, and Jan Peters.
\newblock Motion planning diffusion: Learning and planning of robot motions with diffusion models, 2023.

\bibitem[\protect\citeauthoryear{Chen \bgroup \em et al.\egroup }{2021}]{chen2021decision}
Lili Chen, Kevin Lu, Aravind Rajeswaran, Kimin Lee, Aditya Grover, Misha Laskin, Pieter Abbeel, Aravind Srinivas, and Igor Mordatch.
\newblock Decision transformer: Reinforcement learning via sequence modeling.
\newblock {\em Advances in neural information processing systems}, 34:15084--15097, 2021.

\bibitem[\protect\citeauthoryear{Chen \bgroup \em et al.\egroup }{2022}]{Chen2022ExecutingYC}
Xin Chen, Biao Jiang, Wen Liu, Zilong Huang, Bin Fu, Tao Chen, Jingyi Yu, and Gang Yu.
\newblock Executing your commands via motion diffusion in latent space.
\newblock {\em 2023 IEEE/CVF Conference on Computer Vision and Pattern Recognition (CVPR)}, pages 18000--18010, 2022.

\bibitem[\protect\citeauthoryear{Chen \bgroup \em et al.\egroup }{2023a}]{chen2023offline}
Huayu Chen, Cheng Lu, Chengyang Ying, Hang Su, and Jun Zhu.
\newblock Offline reinforcement learning via high-fidelity generative behavior modeling.
\newblock In {\em The Eleventh International Conference on Learning Representations}, 2023.

\bibitem[\protect\citeauthoryear{Chen \bgroup \em et al.\egroup }{2023b}]{chen2023equidiff}
Kehua Chen, Xianda Chen, Zihan Yu, Meixin Zhu, and Hai Yang.
\newblock Equidiff: A conditional equivariant diffusion model for trajectory prediction, 2023.

\bibitem[\protect\citeauthoryear{Chen \bgroup \em et al.\egroup }{2023c}]{chen2023playfusion}
Lili Chen, Shikhar Bahl, and Deepak Pathak.
\newblock Playfusion: Skill acquisition via diffusion from language-annotated play.
\newblock In {\em 7th Annual Conference on Robot Learning}, 2023.

\bibitem[\protect\citeauthoryear{Chen \bgroup \em et al.\egroup }{2023d}]{chen2023boosting}
Yuhui Chen, Haoran Li, and Dongbin Zhao.
\newblock Boosting continuous control with consistency policy, 2023.

\bibitem[\protect\citeauthoryear{Chen \bgroup \em et al.\egroup }{2023e}]{chen2023genaug}
Zoey Chen, Sho Kiami, Abhishek Gupta, and Vikash Kumar.
\newblock Genaug: Retargeting behaviors to unseen situations via generative augmentation.
\newblock {\em arXiv preprint arXiv:2302.06671}, 2023.

\bibitem[\protect\citeauthoryear{Chi \bgroup \em et al.\egroup }{2023}]{chi2023diffusion}
Cheng Chi, Siyuan Feng, Yilun Du, Zhenjia Xu, Eric Cousineau, Benjamin Burchfiel, and Shuran Song.
\newblock Diffusion policy: Visuomotor policy learning via action diffusion, 2023.

\bibitem[\protect\citeauthoryear{Cho \bgroup \em et al.\egroup }{2022}]{cho2022s2p}
Daesol Cho, Dongseok Shim, and H~Jin Kim.
\newblock S2p: State-conditioned image synthesis for data augmentation in offline reinforcement learning.
\newblock {\em Advances in Neural Information Processing Systems}, 35:11534--11546, 2022.

\bibitem[\protect\citeauthoryear{Dabral \bgroup \em et al.\egroup }{2022}]{Dabral2022MoFusionAF}
Rishabh Dabral, Muhammad~Hamza Mughal, Vladislav Golyanik, and Christian Theobalt.
\newblock Mofusion: A framework for denoising-diffusion-based motion synthesis.
\newblock {\em 2023 IEEE/CVF Conference on Computer Vision and Pattern Recognition (CVPR)}, pages 9760--9770, 2022.

\bibitem[\protect\citeauthoryear{Dhariwal and Nichol}{2021}]{dhariwal2021diffusion}
Prafulla Dhariwal and Alexander Nichol.
\newblock Diffusion models beat gans on image synthesis.
\newblock {\em Advances in neural information processing systems}, 34:8780--8794, 2021.

\bibitem[\protect\citeauthoryear{Dong \bgroup \em et al.\egroup }{2023}]{dong2023aligndiff}
Zibin Dong, Yifu Yuan, Jianye Hao, Fei Ni, Yao Mu, Yan Zheng, Yujing Hu, Tangjie Lv, Changjie Fan, and Zhipeng Hu.
\newblock Aligndiff: Aligning diverse human preferences via behavior-customisable diffusion model, 2023.

\bibitem[\protect\citeauthoryear{Du \bgroup \em et al.\egroup }{2023a}]{du2023learning}
Yilun Du, Mengjiao Yang, Bo~Dai, Hanjun Dai, Ofir Nachum, Joshua~B. Tenenbaum, Dale Schuurmans, and Pieter Abbeel.
\newblock Learning universal policies via text-guided video generation, 2023.

\bibitem[\protect\citeauthoryear{Du \bgroup \em et al.\egroup }{2023b}]{du2023video}
Yilun Du, Mengjiao Yang, Pete Florence, Fei Xia, Ayzaan Wahid, Brian Ichter, Pierre Sermanet, Tianhe Yu, Pieter Abbeel, Joshua~B. Tenenbaum, Leslie Kaelbling, Andy Zeng, and Jonathan Tompson.
\newblock Video language planning, 2023.

\bibitem[\protect\citeauthoryear{Fujimoto and Gu}{2021}]{fujimoto2021minimalist}
Scott Fujimoto and Shixiang~Shane Gu.
\newblock A minimalist approach to offline reinforcement learning.
\newblock {\em Advances in neural information processing systems}, 34:20132--20145, 2021.

\bibitem[\protect\citeauthoryear{Fujimoto \bgroup \em et al.\egroup }{2019}]{fujimoto2019off}
Scott Fujimoto, David Meger, and Doina Precup.
\newblock Off-policy deep reinforcement learning without exploration.
\newblock In {\em International conference on machine learning}, pages 2052--2062. PMLR, 2019.

\bibitem[\protect\citeauthoryear{Gao \bgroup \em et al.\egroup }{2023}]{gao2023pretrained}
Jialu Gao, Kaizhe Hu, Guowei Xu, and Huazhe Xu.
\newblock Can pre-trained text-to-image models generate visual goals for reinforcement learning?, 2023.

\bibitem[\protect\citeauthoryear{Goodfellow \bgroup \em et al.\egroup }{2014}]{goodfellow2014generative}
Ian Goodfellow, Jean Pouget-Abadie, Mehdi Mirza, Bing Xu, David Warde-Farley, Sherjil Ozair, Aaron Courville, and Yoshua Bengio.
\newblock Generative adversarial nets.
\newblock {\em Advances in neural information processing systems}, 27, 2014.

\bibitem[\protect\citeauthoryear{Ha \bgroup \em et al.\egroup }{2023}]{ha2023scaling}
Huy Ha, Pete Florence, and Shuran Song.
\newblock Scaling up and distilling down: Language-guided robot skill acquisition.
\newblock In {\em 7th Annual Conference on Robot Learning}, 2023.

\bibitem[\protect\citeauthoryear{Hansen-Estruch \bgroup \em et al.\egroup }{2023}]{hansenestruch2023idql}
Philippe Hansen-Estruch, Ilya Kostrikov, Michael Janner, Jakub~Grudzien Kuba, and Sergey Levine.
\newblock Idql: Implicit q-learning as an actor-critic method with diffusion policies, 2023.

\bibitem[\protect\citeauthoryear{He \bgroup \em et al.\egroup }{2023a}]{he2023diffusion}
Haoran He, Chenjia Bai, Kang Xu, Zhuoran Yang, Weinan Zhang, Dong Wang, Bin Zhao, and Xuelong Li.
\newblock Diffusion model is an effective planner and data synthesizer for multi-task reinforcement learning.
\newblock 2023.

\bibitem[\protect\citeauthoryear{He \bgroup \em et al.\egroup }{2023b}]{he2023diffcps}
Longxiang He, Linrui Zhang, Junbo Tan, and Xueqian Wang.
\newblock Diffcps: Diffusion model based constrained policy search for offline reinforcement learning, 2023.

\bibitem[\protect\citeauthoryear{Hegde \bgroup \em et al.\egroup }{2023}]{hegde2023generating}
Shashank Hegde, Sumeet Batra, K.~R. Zentner, and Gaurav~S. Sukhatme.
\newblock Generating behaviorally diverse policies with latent diffusion models, 2023.

\bibitem[\protect\citeauthoryear{Ho and Salimans}{2022}]{ho2022classifierfree}
Jonathan Ho and Tim Salimans.
\newblock Classifier-free diffusion guidance, 2022.

\bibitem[\protect\citeauthoryear{Ho \bgroup \em et al.\egroup }{2020}]{ho2020denoising}
Jonathan Ho, Ajay Jain, and Pieter Abbeel.
\newblock Denoising diffusion probabilistic models, 2020.

\bibitem[\protect\citeauthoryear{Hu \bgroup \em et al.\egroup }{2023}]{hu2023instructed}
Jifeng Hu, Yanchao Sun, Sili Huang, SiYuan Guo, Hechang Chen, Li~Shen, Lichao Sun, Yi~Chang, and Dacheng Tao.
\newblock Instructed diffuser with temporal condition guidance for offline reinforcement learning, 2023.

\bibitem[\protect\citeauthoryear{Imre}{2021}]{imre2021investigation}
Baris Imre.
\newblock An investigation of generative replay in deep reinforcement learning.
\newblock {B.S.} thesis, University of Twente, 2021.

\bibitem[\protect\citeauthoryear{Janner \bgroup \em et al.\egroup }{2022}]{Janner2022PlanningWD}
Michael Janner, Yilun Du, Joshua~B. Tenenbaum, and Sergey Levine.
\newblock Planning with diffusion for flexible behavior synthesis.
\newblock In {\em International Conference on Machine Learning}, 2022.

\bibitem[\protect\citeauthoryear{Jiang \bgroup \em et al.\egroup }{2023}]{jiang2023motiondiffuser}
Chiyu~Max Jiang, Andre Cornman, Cheolho Park, Ben Sapp, Yin Zhou, and Dragomir Anguelov.
\newblock Motiondiffuser: Controllable multi-agent motion prediction using diffusion, 2023.

\bibitem[\protect\citeauthoryear{Kandpal \bgroup \em et al.\egroup }{2023}]{kandpal2023large}
Nikhil Kandpal, Haikang Deng, Adam Roberts, Eric Wallace, and Colin Raffel.
\newblock Large language models struggle to learn long-tail knowledge.
\newblock In {\em International Conference on Machine Learning}, pages 15696--15707. PMLR, 2023.

\bibitem[\protect\citeauthoryear{Kang \bgroup \em et al.\egroup }{2023}]{kang2023efficient}
Bingyi Kang, Xiao Ma, Chao Du, Tianyu Pang, and Shuicheng Yan.
\newblock Efficient diffusion policies for offline reinforcement learning, 2023.

\bibitem[\protect\citeauthoryear{Kapelyukh \bgroup \em et al.\egroup }{2022}]{Kapelyukh2022DALLEBotIW}
Ivan Kapelyukh, Vitalis Vosylius, and Edward Johns.
\newblock Dall-e-bot: Introducing web-scale diffusion models to robotics.
\newblock {\em IEEE Robotics and Automation Letters}, 8:3956--3963, 2022.

\bibitem[\protect\citeauthoryear{Katara \bgroup \em et al.\egroup }{2023}]{katara2023gen2sim}
Pushkal Katara, Zhou Xian, and Katerina Fragkiadaki.
\newblock Gen2sim: Scaling up robot learning in simulation with generative models.
\newblock {\em arXiv preprint arXiv:2310.18308}, 2023.

\bibitem[\protect\citeauthoryear{Kingma and Welling}{2013}]{kingma2013auto}
Diederik~P Kingma and Max Welling.
\newblock Auto-encoding variational bayes.
\newblock {\em arXiv preprint arXiv:1312.6114}, 2013.

\bibitem[\protect\citeauthoryear{Kiran \bgroup \em et al.\egroup }{2021}]{kiran2021deep}
B~Ravi Kiran, Ibrahim Sobh, Victor Talpaert, Patrick Mannion, Ahmad~A Al~Sallab, Senthil Yogamani, and Patrick P{\'e}rez.
\newblock Deep reinforcement learning for autonomous driving: A survey.
\newblock {\em IEEE Transactions on Intelligent Transportation Systems}, 23(6):4909--4926, 2021.

\bibitem[\protect\citeauthoryear{Ko \bgroup \em et al.\egroup }{2023}]{ko2023learning}
Po-Chen Ko, Jiayuan Mao, Yilun Du, Shao-Hua Sun, and Joshua~B. Tenenbaum.
\newblock Learning to act from actionless videos through dense correspondences, 2023.

\bibitem[\protect\citeauthoryear{Kober \bgroup \em et al.\egroup }{2013}]{kober2013reinforcement}
Jens Kober, J~Andrew Bagnell, and Jan Peters.
\newblock Reinforcement learning in robotics: A survey.
\newblock {\em The International Journal of Robotics Research}, 32(11):1238--1274, 2013.

\bibitem[\protect\citeauthoryear{Kong \bgroup \em et al.\egroup }{2020}]{kong2020diffwave}
Zhifeng Kong, Wei Ping, Jiaji Huang, Kexin Zhao, and Bryan Catanzaro.
\newblock Diffwave: A versatile diffusion model for audio synthesis.
\newblock {\em arXiv preprint arXiv:2009.09761}, 2020.

\bibitem[\protect\citeauthoryear{Kostrikov \bgroup \em et al.\egroup }{2021}]{kostrikov2021offline}
Ilya Kostrikov, Rob Fergus, Jonathan Tompson, and Ofir Nachum.
\newblock Offline reinforcement learning with fisher divergence critic regularization.
\newblock In {\em International Conference on Machine Learning}, pages 5774--5783. PMLR, 2021.

\bibitem[\protect\citeauthoryear{Kumar \bgroup \em et al.\egroup }{2020}]{kumar2020conservative}
Aviral Kumar, Aurick Zhou, George Tucker, and Sergey Levine.
\newblock Conservative q-learning for offline reinforcement learning.
\newblock {\em Advances in Neural Information Processing Systems}, 33:1179--1191, 2020.

\bibitem[\protect\citeauthoryear{Laskin \bgroup \em et al.\egroup }{2020}]{laskin2020reinforcement}
Misha Laskin, Kimin Lee, Adam Stooke, Lerrel Pinto, Pieter Abbeel, and Aravind Srinivas.
\newblock Reinforcement learning with augmented data.
\newblock {\em Advances in neural information processing systems}, 33:19884--19895, 2020.

\bibitem[\protect\citeauthoryear{Lee and Han}{2021}]{lee2021nu}
Junhyeok Lee and Seungu Han.
\newblock Nu-wave: A diffusion probabilistic model for neural audio upsampling.
\newblock {\em arXiv preprint arXiv:2104.02321}, 2021.

\bibitem[\protect\citeauthoryear{Li \bgroup \em et al.\egroup }{2022}]{li2022diffusion}
Xiang Li, John Thickstun, Ishaan Gulrajani, Percy~S Liang, and Tatsunori~B Hashimoto.
\newblock Diffusion-lm improves controllable text generation.
\newblock {\em Advances in Neural Information Processing Systems}, 35:4328--4343, 2022.

\bibitem[\protect\citeauthoryear{Li \bgroup \em et al.\egroup }{2023a}]{Hier_diffusion}
Wenhao Li, Xiangfeng Wang, Bo~Jin, and Hongyuan Zha.
\newblock Hierarchical diffusion for offline decision making.
\newblock In Andreas Krause, Emma Brunskill, Kyunghyun Cho, Barbara Engelhardt, Sivan Sabato, and Jonathan Scarlett, editors, {\em Proceedings of the 40th International Conference on Machine Learning}, volume 202 of {\em Proceedings of Machine Learning Research}, pages 20035--20064. PMLR, 23--29 Jul 2023.

\bibitem[\protect\citeauthoryear{Li \bgroup \em et al.\egroup }{2023b}]{li2023crossway}
Xiang Li, Varun Belagali, Jinghuan Shang, and Michael~S. Ryoo.
\newblock Crossway diffusion: Improving diffusion-based visuomotor policy via self-supervised learning, 2023.

\bibitem[\protect\citeauthoryear{Li \bgroup \em et al.\egroup }{2023c}]{li2023conservatism}
Zhuoran Li, Ling Pan, and Longbo Huang.
\newblock Beyond conservatism: Diffusion policies in offline multi-agent reinforcement learning, 2023.

\bibitem[\protect\citeauthoryear{Li}{2023}]{li2023efficient}
Wenhao Li.
\newblock Efficient planning with latent diffusion, 2023.

\bibitem[\protect\citeauthoryear{Liang \bgroup \em et al.\egroup }{2023a}]{adaptdiffuser}
Zhixuan Liang, Yao Mu, Mingyu Ding, Fei Ni, Masayoshi Tomizuka, and Ping Luo.
\newblock {A}dapt{D}iffuser: Diffusion models as adaptive self-evolving planners.
\newblock In Andreas Krause, Emma Brunskill, Kyunghyun Cho, Barbara Engelhardt, Sivan Sabato, and Jonathan Scarlett, editors, {\em Proceedings of the 40th International Conference on Machine Learning}, volume 202 of {\em Proceedings of Machine Learning Research}, pages 20725--20745. PMLR, 23--29 Jul 2023.

\bibitem[\protect\citeauthoryear{Liang \bgroup \em et al.\egroup }{2023b}]{liang2023skilldiffuser}
Zhixuan Liang, Yao Mu, Hengbo Ma, Masayoshi Tomizuka, Mingyu Ding, and Ping Luo.
\newblock Skilldiffuser: Interpretable hierarchical planning via skill abstractions in diffusion-based task execution.
\newblock {\em arXiv preprint arXiv:2312.11598}, 2023.

\bibitem[\protect\citeauthoryear{Liu \bgroup \em et al.\egroup }{2021}]{liu2021conflict}
Bo~Liu, Xingchao Liu, Xiaojie Jin, Peter Stone, and Qiang Liu.
\newblock Conflict-averse gradient descent for multi-task learning.
\newblock {\em Advances in Neural Information Processing Systems}, 34:18878--18890, 2021.

\bibitem[\protect\citeauthoryear{Liu \bgroup \em et al.\egroup }{2023}]{liu2023constrained}
Zuxin Liu, Zijian Guo, Yihang Yao, Zhepeng Cen, Wenhao Yu, Tingnan Zhang, and Ding Zhao.
\newblock Constrained decision transformer for offline safe reinforcement learning.
\newblock {\em arXiv preprint arXiv:2302.07351}, 2023.

\bibitem[\protect\citeauthoryear{Lu \bgroup \em et al.\egroup }{2022}]{lu2022dpmsolver}
Cheng Lu, Yuhao Zhou, Fan Bao, Jianfei Chen, Chongxuan Li, and Jun Zhu.
\newblock Dpm-solver: A fast ode solver for diffusion probabilistic model sampling in around 10 steps, 2022.

\bibitem[\protect\citeauthoryear{Lu \bgroup \em et al.\egroup }{2023a}]{lu2023contrastive}
Cheng Lu, Huayu Chen, Jianfei Chen, Hang Su, Chongxuan Li, and Jun Zhu.
\newblock Contrastive energy prediction for exact energy-guided diffusion sampling in offline reinforcement learning, 2023.

\bibitem[\protect\citeauthoryear{Lu \bgroup \em et al.\egroup }{2023b}]{lu2023synthetic}
Cong Lu, Philip~J. Ball, and Jack Parker-Holder.
\newblock Synthetic experience replay.
\newblock In {\em Workshop on Reincarnating Reinforcement Learning at ICLR 2023}, 2023.

\bibitem[\protect\citeauthoryear{Lugmayr \bgroup \em et al.\egroup }{2022}]{lugmayr2022repaint}
Andreas Lugmayr, Martin Danelljan, Andres Romero, Fisher Yu, Radu Timofte, and Luc Van~Gool.
\newblock Repaint: Inpainting using denoising diffusion probabilistic models.
\newblock In {\em Proceedings of the IEEE/CVF Conference on Computer Vision and Pattern Recognition}, pages 11461--11471, 2022.

\bibitem[\protect\citeauthoryear{Luo and Hu}{2021}]{luo2021diffusion}
Shitong Luo and Wei Hu.
\newblock Diffusion probabilistic models for 3d point cloud generation.
\newblock In {\em Proceedings of the IEEE/CVF Conference on Computer Vision and Pattern Recognition}, pages 2837--2845, 2021.

\bibitem[\protect\citeauthoryear{Luo \bgroup \em et al.\egroup }{2022}]{luo2022survey}
Fan-Ming Luo, Tian Xu, Hang Lai, Xiong-Hui Chen, Weinan Zhang, and Yang Yu.
\newblock A survey on model-based reinforcement learning.
\newblock {\em arXiv preprint arXiv:2206.09328}, 2022.

\bibitem[\protect\citeauthoryear{Lyu \bgroup \em et al.\egroup }{2022}]{lyu2022mildly}
Jiafei Lyu, Xiaoteng Ma, Xiu Li, and Zongqing Lu.
\newblock Mildly conservative q-learning for offline reinforcement learning.
\newblock {\em Advances in Neural Information Processing Systems}, 35:1711--1724, 2022.

\bibitem[\protect\citeauthoryear{Mazoure \bgroup \em et al.\egroup }{2023}]{mazoure2023value}
Bogdan Mazoure, Walter Talbott, Miguel~Angel Bautista, Devon Hjelm, Alexander Toshev, and Josh Susskind.
\newblock Value function estimation using conditional diffusion models for control, 2023.

\bibitem[\protect\citeauthoryear{Mishra \bgroup \em et al.\egroup }{2023}]{mishra2023generative}
Utkarsh~Aashu Mishra, Shangjie Xue, Yongxin Chen, and Danfei Xu.
\newblock Generative skill chaining: Long-horizon skill planning with diffusion models.
\newblock In {\em 7th Annual Conference on Robot Learning}, 2023.

\bibitem[\protect\citeauthoryear{Mnih \bgroup \em et al.\egroup }{2013}]{mnih2013playing}
Volodymyr Mnih, Koray Kavukcuoglu, David Silver, Alex Graves, Ioannis Antonoglou, Daan Wierstra, and Martin Riedmiller.
\newblock Playing atari with deep reinforcement learning.
\newblock {\em arXiv preprint arXiv:1312.5602}, 2013.

\bibitem[\protect\citeauthoryear{Nagabandi \bgroup \em et al.\egroup }{2018}]{nagabandi2018neural}
Anusha Nagabandi, Gregory Kahn, Ronald~S Fearing, and Sergey Levine.
\newblock Neural network dynamics for model-based deep reinforcement learning with model-free fine-tuning.
\newblock In {\em 2018 IEEE international conference on robotics and automation (ICRA)}, pages 7559--7566. IEEE, 2018.

\bibitem[\protect\citeauthoryear{Neal and others}{2011}]{neal2011mcmc}
Radford~M Neal et~al.
\newblock Mcmc using hamiltonian dynamics.
\newblock {\em Handbook of markov chain monte carlo}, 2(11):2, 2011.

\bibitem[\protect\citeauthoryear{Ng \bgroup \em et al.\egroup }{2023}]{ng2023diffusion}
Eley Ng, Ziang Liu, and Monroe Kennedy~III au2.
\newblock Diffusion co-policy for synergistic human-robot collaborative tasks, 2023.

\bibitem[\protect\citeauthoryear{Ni \bgroup \em et al.\egroup }{2023}]{ni2023metadiffuser}
Fei Ni, Jianye Hao, Yao Mu, Yifu Yuan, Yan Zheng, Bin Wang, and Zhixuan Liang.
\newblock Metadiffuser: Diffusion model as conditional planner for offline meta-rl, 2023.

\bibitem[\protect\citeauthoryear{Pearce \bgroup \em et al.\egroup }{2023}]{pearce2023imitating}
Tim Pearce, Tabish Rashid, Anssi Kanervisto, Dave Bignell, Mingfei Sun, Raluca Georgescu, Sergio~Valcarcel Macua, Shan~Zheng Tan, Ida Momennejad, Katja Hofmann, and Sam Devlin.
\newblock Imitating human behaviour with diffusion models.
\newblock In {\em The Eleventh International Conference on Learning Representations}, 2023.

\bibitem[\protect\citeauthoryear{Qin \bgroup \em et al.\egroup }{2020}]{qin2020user}
Jiarui Qin, Weinan Zhang, Xin Wu, Jiarui Jin, Yuchen Fang, and Yong Yu.
\newblock User behavior retrieval for click-through rate prediction.
\newblock In {\em Proceedings of the 43rd International ACM SIGIR Conference on Research and Development in Information Retrieval}, pages 2347--2356, 2020.

\bibitem[\protect\citeauthoryear{Raab \bgroup \em et al.\egroup }{2023}]{raab2023single}
Sigal Raab, Inbal Leibovitch, Guy Tevet, Moab Arar, Amit~H. Bermano, and Daniel Cohen-Or.
\newblock Single motion diffusion, 2023.

\bibitem[\protect\citeauthoryear{Reed \bgroup \em et al.\egroup }{2022}]{reed2022generalist}
Scott Reed, Konrad Zolna, Emilio Parisotto, Sergio~Gomez Colmenarejo, Alexander Novikov, Gabriel Barth-Maron, Mai Gimenez, Yury Sulsky, Jackie Kay, Jost~Tobias Springenberg, et~al.
\newblock A generalist agent.
\newblock {\em arXiv preprint arXiv:2205.06175}, 2022.

\bibitem[\protect\citeauthoryear{Reuss \bgroup \em et al.\egroup }{2023}]{reuss2023goalconditioned}
Moritz Reuss, Maximilian Li, Xiaogang Jia, and Rudolf Lioutikov.
\newblock Goal-conditioned imitation learning using score-based diffusion policies, 2023.

\bibitem[\protect\citeauthoryear{Rigter \bgroup \em et al.\egroup }{2023}]{rigter2023world}
Marc Rigter, Jun Yamada, and Ingmar Posner.
\newblock World models via policy-guided trajectory diffusion.
\newblock {\em arXiv preprint arXiv:2312.08533}, 2023.

\bibitem[\protect\citeauthoryear{Schmidhuber}{2019}]{schmidhuber2019reinforcement}
Juergen Schmidhuber.
\newblock Reinforcement learning upside down: Don't predict rewards--just map them to actions.
\newblock {\em arXiv preprint arXiv:1912.02875}, 2019.

\bibitem[\protect\citeauthoryear{Schneuing \bgroup \em et al.\egroup }{2022}]{schneuing2022structure}
Arne Schneuing, Yuanqi Du, Charles Harris, Arian Jamasb, Ilia Igashov, Weitao Du, Tom Blundell, Pietro Li{\'o}, Carla Gomes, Max Welling, et~al.
\newblock Structure-based drug design with dvariant diffusion models.
\newblock {\em arXiv preprint arXiv:2210.13695}, 2022.

\bibitem[\protect\citeauthoryear{Schrittwieser \bgroup \em et al.\egroup }{2020}]{schrittwieser2020mastering}
Julian Schrittwieser, Ioannis Antonoglou, Thomas Hubert, Karen Simonyan, Laurent Sifre, Simon Schmitt, Arthur Guez, Edward Lockhart, Demis Hassabis, Thore Graepel, et~al.
\newblock Mastering atari, go, chess and shogi by planning with a learned model.
\newblock {\em Nature}, 588(7839):604--609, 2020.

\bibitem[\protect\citeauthoryear{Sheynin \bgroup \em et al.\egroup }{2022}]{sheynin2022knn}
Shelly Sheynin, Oron Ashual, Adam Polyak, Uriel Singer, Oran Gafni, Eliya Nachmani, and Yaniv Taigman.
\newblock Knn-diffusion: Image generation via large-scale retrieval.
\newblock {\em arXiv preprint arXiv:2204.02849}, 2022.

\bibitem[\protect\citeauthoryear{Sinha \bgroup \em et al.\egroup }{2021}]{sinha2021s4rl}
Samarth Sinha, Ajay Mandlekar, and Animesh Garg.
\newblock S4rl: Surprisingly simple self-supervision for offline reinforcement learning, 2021.

\bibitem[\protect\citeauthoryear{Song \bgroup \em et al.\egroup }{2021}]{song2021scorebased}
Yang Song, Jascha Sohl-Dickstein, Diederik~P. Kingma, Abhishek Kumar, Stefano Ermon, and Ben Poole.
\newblock Score-based generative modeling through stochastic differential equations, 2021.

\bibitem[\protect\citeauthoryear{Song \bgroup \em et al.\egroup }{2022}]{song2022denoising}
Jiaming Song, Chenlin Meng, and Stefano Ermon.
\newblock Denoising diffusion implicit models, 2022.

\bibitem[\protect\citeauthoryear{Sridhar \bgroup \em et al.\egroup }{2023}]{sridhar2023nomad}
Ajay Sridhar, Dhruv Shah, Catherine Glossop, and Sergey Levine.
\newblock Nomad: Goal masked diffusion policies for navigation and exploration, 2023.

\bibitem[\protect\citeauthoryear{Sutton and Barto}{2018}]{sutton2018reinforcement}
Richard~S Sutton and Andrew~G Barto.
\newblock {\em Reinforcement learning: An introduction}.
\newblock MIT press, 2018.

\bibitem[\protect\citeauthoryear{Taiga \bgroup \em et al.\egroup }{2022}]{taiga2022investigating}
Adrien~Ali Taiga, Rishabh Agarwal, Jesse Farebrother, Aaron Courville, and Marc~G Bellemare.
\newblock Investigating multi-task pretraining and generalization in reinforcement learning.
\newblock In {\em The Eleventh International Conference on Learning Representations}, 2022.

\bibitem[\protect\citeauthoryear{Tevet \bgroup \em et al.\egroup }{2022}]{tevet2022humandiffuse}
Guy Tevet, Sigal Raab, Brian Gordon, Yonatan Shafir, Daniel Cohen-Or, and Amit~H. Bermano.
\newblock Human motion diffusion model, 2022.

\bibitem[\protect\citeauthoryear{Venkatraman \bgroup \em et al.\egroup }{2023}]{venkatraman2023reasoning}
Siddarth Venkatraman, Shivesh Khaitan, Ravi~Tej Akella, John Dolan, Jeff Schneider, and Glen Berseth.
\newblock Reasoning with latent diffusion in offline reinforcement learning, 2023.

\bibitem[\protect\citeauthoryear{Vithayathil~Varghese and Mahmoud}{2020}]{vithayathil2020survey}
Nelson Vithayathil~Varghese and Qusay~H Mahmoud.
\newblock A survey of multi-task deep reinforcement learning.
\newblock {\em Electronics}, 9(9):1363, 2020.

\bibitem[\protect\citeauthoryear{Wang \bgroup \em et al.\egroup }{2023}]{wang2023diffusion}
Zhendong Wang, Jonathan~J Hunt, and Mingyuan Zhou.
\newblock Diffusion policies as an expressive policy class for offline reinforcement learning.
\newblock In {\em The Eleventh International Conference on Learning Representations}, 2023.

\bibitem[\protect\citeauthoryear{Watson \bgroup \em et al.\egroup }{2021}]{watson2021learning}
Daniel Watson, Jonathan Ho, Mohammad Norouzi, and William Chan.
\newblock Learning to efficiently sample from diffusion probabilistic models, 2021.

\bibitem[\protect\citeauthoryear{Xian \bgroup \em et al.\egroup }{2023}]{xian2023chaineddiffuser}
Zhou Xian, Nikolaos Gkanatsios, Theophile Gervet, Tsung-Wei Ke, and Katerina Fragkiadaki.
\newblock Chaineddiffuser: Unifying trajectory diffusion and keypose prediction for robotic manipulation.
\newblock In {\em 7th Annual Conference on Robot Learning}, 2023.

\bibitem[\protect\citeauthoryear{Xiao \bgroup \em et al.\egroup }{2019}]{xiao2019learning}
Chenjun Xiao, Yifan Wu, Chen Ma, Dale Schuurmans, and Martin M{\"u}ller.
\newblock Learning to combat compounding-error in model-based reinforcement learning.
\newblock {\em arXiv preprint arXiv:1912.11206}, 2019.

\bibitem[\protect\citeauthoryear{Xiao \bgroup \em et al.\egroup }{2023}]{xiao2023safediffuser}
Wei Xiao, Tsun-Hsuan Wang, Chuang Gan, and Daniela Rus.
\newblock Safediffuser: Safe planning with diffusion probabilistic models, 2023.

\bibitem[\protect\citeauthoryear{Xu \bgroup \em et al.\egroup }{2022}]{xu2022geodiff}
Minkai Xu, Lantao Yu, Yang Song, Chence Shi, Stefano Ermon, and Jian Tang.
\newblock Geodiff: A geometric diffusion model for molecular conformation generation.
\newblock {\em arXiv preprint arXiv:2203.02923}, 2022.

\bibitem[\protect\citeauthoryear{Xu \bgroup \em et al.\egroup }{2023}]{xu2023xskill}
Mengda Xu, Zhenjia Xu, Cheng Chi, Manuela Veloso, and Shuran Song.
\newblock Xskill: Cross embodiment skill discovery, 2023.

\bibitem[\protect\citeauthoryear{Yang \bgroup \em et al.\egroup }{2023a}]{yang2023policy}
Long Yang, Zhixiong Huang, Fenghao Lei, Yucun Zhong, Yiming Yang, Cong Fang, Shiting Wen, Binbin Zhou, and Zhouchen Lin.
\newblock Policy representation via diffusion probability model for reinforcement learning.
\newblock {\em arXiv preprint arXiv:2305.13122}, 2023.

\bibitem[\protect\citeauthoryear{Yang \bgroup \em et al.\egroup }{2023b}]{yang2023learning}
Mengjiao Yang, Yilun Du, Kamyar Ghasemipour, Jonathan Tompson, Dale Schuurmans, and Pieter Abbeel.
\newblock Learning interactive real-world simulators, 2023.

\bibitem[\protect\citeauthoryear{Yoneda \bgroup \em et al.\egroup }{2023}]{yoneda2023noise}
Takuma Yoneda, Luzhe Sun, , Ge~Yang, Bradly Stadie, and Matthew Walter.
\newblock To the noise and back: Diffusion for shared autonomy, 2023.

\bibitem[\protect\citeauthoryear{Yu \bgroup \em et al.\egroup }{2023}]{yu2023scaling}
Tianhe Yu, Ted Xiao, Austin Stone, Jonathan Tompson, Anthony Brohan, Su~Wang, Jaspiar Singh, Clayton Tan, Jodilyn Peralta, Brian Ichter, et~al.
\newblock Scaling robot learning with semantically imagined experience.
\newblock {\em arXiv preprint arXiv:2302.11550}, 2023.

\bibitem[\protect\citeauthoryear{Zhang \bgroup \em et al.\egroup }{2022}]{zhang2022motiondiffuse}
Mingyuan Zhang, Zhongang Cai, Liang Pan, Fangzhou Hong, Xinying Guo, Lei Yang, and Ziwei Liu.
\newblock Motiondiffuse: Text-driven human motion generation with diffusion model.
\newblock {\em arXiv preprint arXiv:2208.15001}, 2022.

\bibitem[\protect\citeauthoryear{Zhang \bgroup \em et al.\egroup }{2023a}]{language-control-diffusion}
Edwin Zhang, Yujie Lu, William Wang, and Amy Zhang.
\newblock Lad: Language control diffusion: efficiently scaling through space, time, and tasks.
\newblock {\em arXiv preprint arXiv:2210.15629}, 2023.

\bibitem[\protect\citeauthoryear{Zhang \bgroup \em et al.\egroup }{2023b}]{zhang2023remodiffuse}
Mingyuan Zhang, Xinying Guo, Liang Pan, Zhongang Cai, Fangzhou Hong, Huirong Li, Lei Yang, and Ziwei Liu.
\newblock Remodiffuse: Retrieval-augmented motion diffusion model, 2023.

\bibitem[\protect\citeauthoryear{Zheng \bgroup \em et al.\egroup }{2023}]{zheng2023truncated}
Huangjie Zheng, Pengcheng He, Weizhu Chen, and Mingyuan Zhou.
\newblock Truncated diffusion probabilistic models and diffusion-based adversarial auto-encoders, 2023.

\bibitem[\protect\citeauthoryear{Zhu \bgroup \em et al.\egroup }{2021}]{zhu2021mapgo}
Menghui Zhu, Minghuan Liu, Jian Shen, Zhicheng Zhang, Sheng Chen, Weinan Zhang, Deheng Ye, Yong Yu, Qiang Fu, and Wei Yang.
\newblock Mapgo: Model-assisted policy optimization for goal-oriented tasks.
\newblock {\em arXiv preprint arXiv:2105.06350}, 2021.

\bibitem[\protect\citeauthoryear{Zhu \bgroup \em et al.\egroup }{2023}]{zhu2023madiff}
Zhengbang Zhu, Minghuan Liu, Liyuan Mao, Bingyi Kang, Minkai Xu, Yong Yu, Stefano Ermon, and Weinan Zhang.
\newblock Madiff: Offline multi-agent learning with diffusion models, 2023.

\end{thebibliography}

\end{document}